# Exploring Novel Pooling Strategies for Edge Preserved Feature Maps in Convolutional Neural Networks

Adithya Sineesh [a, *] and Mahesh Raveendranatha Panicker [b, *, **]

[a] National Institute of Technology Trichy, Tamil Nadu, India

[b] Indian Institute of Technology Palakkad, Kerala, India

**Abstract**

With the introduction of anti-aliased convolutional neural networks (CNN), there has been some resurgence in relooking the way pooling is done in CNNs. The fundamental building block of the anti-aliased CNN has been the application of Gaussian smoothing before the pooling operation to reduce the distortion due to aliasing thereby making CNNs shift invariant. Wavelet based approaches have also been proposed as a possibility of additional noise removal capability and gave interesting results for even segmentation tasks. However, all the approaches proposed completely remove the high frequency components under the assumption that they are noise. However, by removing high frequency components, the edges in the feature maps are also smoothed. In this work, an exhaustive analysis of the edge preserving pooling options for classification, segmentation and autoencoders are presented. Two novel pooling approaches are presented such as Laplacian-Gaussian Concatenation with Attention (LGCA) pooling and Wavelet based approximate-detailed coefficient concatenation with attention (WADCA) pooling. The results suggest that the proposed pooling approaches outperform the conventional pooling as well as blur pooling for classification, segmentation and autoencoders. In terms of average binary classification accuracy (cats vs dogs), the proposed LGCA approach outperforms the normal pooling and blur pooling by 4% and 2%, 3% and 4%, 3% and 0.5% for MobileNetv2,

DenseNet121 and ResNet50 respectively. On the other hand, the proposed WADCA approach outperforms the normal pooling and blur pooling by 5% and 3%, 2% and 3%, 2 and 0.17% for MobileNetv2, DenseNet121 and ResNet50 respectively. It is also observed from results that edge preserving pooling doesn't have any significance in segmentation tasks possibly due to high resolution to low resolution translation, whereas for convolutional auto encoders high resolution reconstruction has been observed for the LGCA pooling.

## 1. Introduction

One of the unique capabilities of convolutional neural networks (CNNs) has been its ability to extract local and global features while preserving the local neighborhood relationship intact through the processes of convolution and pooling/strides. However, direct downsizing of the features at the output of convolutional layers could result in the aliasing induced distortion [1, 2]. Anti-aliased convolutional neural networks (ACNNs) have been introduced as an attempt towards reducing the distortion effects due to anti-aliasing [1]. As reported in [1], even though max-pooling [3] has been the de facto pooling in most of the modern deep neural networks (VGG, ResNets, DenseNets etc.), the issue has been a rather unstable performance even for small affine transformations of the input images. In [2], it has also been reported that the high frequency random noise gets aliased in the case of max pooling or strided convolutions resulting in weak noise-robustness [4]. To improve the noise agnostic performance of the CNNs, a wavelet based denoising scheme is proposed [2] in the pooling stage. This ensures that no max operation and usage of well researched wavelet basis instead of empirically designed basis in [1].

Despite the recent surge of interest, feature preserving pooling is an area which is not well researched in literature. In [1, 2] and related literature, the emphasis has been on preserving the low frequency features by avoiding aliasing and reducing noise from high frequency regions. However, the main issue with these approaches is the edges in the features will also get smoothed and could adversely affect the performance. In this work, two novel approaches are presented as below:

1. The Laplacian and Gaussian filtered convolutional feature maps are concatenated and weighted by an attention layer, followed by respective pooling. This makes the entire selection of Laplacian or Gaussian filtered kernels learnable as per the application.
2. A wavelet based approach where the convolutional layer output feature is passed through a discrete wavelet transform (DWT) followed by individual reconstruction, attention weighting and respective pooling.

An exhaustive analysis of the above pooling schemes including the conventional max pooling and the blurpool approach in [1] for tasks such as classifiers (ResNet50, Densenet121 and MobileNet), typical convolutional autoencoder and segmentation networks (U-Net and SegNet) is also presented in this work.

**1.1 Related Work**

One of the serious concerns about neural networks in general, and CNNs in particular, in the recent past has been its susceptibility to subtle changes in the form of shifts [1], rotations [5], noise [6], blurring [4] and security attacks [7]. Even Though many of these subtle changes are not perceivable to humans, the performance of the CNNs have been seriously compromised as shown in [1, 2]. There have been many attempts in general to make neural networks more robust against the above issues [ 8-16]. In [8], it was observed that the intermediate layers of a trained network have excessively high frequencies, affecting its shift invariance. The authors recommend using a skip generator and a residual discriminator, without progressive growth to overcome these issues. Xie, Qizhe, et al [9] presents a self-training method where two EfficientNet models are trained on labelled ImageNet images. The first model (teacher) then generates pseudo labels, and the second model (student) is trained on the combination of labelled and pseudo labelled images after injecting noise, making the student able to generalize better than the teacher. In [10], Wang, Sheng-Yu, et al shows that it is possible to identify real images to those generated by a CNN. This is because of artifacts present in common CNN designs that reduce their representational power. E.g.: strided convolutions that reduce translation invariance. To augment CNN with non-local interactions, self-attention has been adopted in [11], where a position-sensitive axial-attention on image classification and segmentation tasks have been discussed in depth. Taori, Rohan, et al. [12] evaluated 204 ImageNet models in 213 different test conditions and the results showed that the natural distribution shifts in data are still an open problem. The authors further claimed that most of the progress has instead been made in synthetic distribution of data. In [13], Ridnik, Tal, et. al. introduces TResNet, a new family of GPU dedicated models with better efficiency and accuracy than ResNet-50. To improve the shift equivariance and robustness of the network, all the downscaling layers are replaced by an equivalent anti-aliasing component as in [1]. Schölkopf, Bernhard, et al in [14] puts forth the argument that causality can contribute towards resolving some of the limitations of machine learning like lack of robustness and inability to learn reusable mechanisms in an expedited manner. In [15], it is shown that modern networks can exploit absolute spatial location all over the image as boundary effects operate even far from the image boundary due to modern CNN filters' huge receptive field. This in turn makes translation invariant. However, this mandates the deep neural network which are generally data and power hungry.

In [1, 2], it is observed that down sampling in the form of max/min/average pooling which sort of dilates the images are the main culprits behind the very susceptible nature of CNNs. Although there have been some attempts to have a more robust pooling operation by employing mixed pooling [16] or stochastic pooling [17], they are just not as common as max pooling and may appear to be just a one-off solution. As discussed before, we have done an exhaustive analysis of the effect of pooling layers in popular classifiers, convolutional autoencoder and segmentation networks.

## 2. Proposed Approaches

As discussed, all the approaches proposed in literature completely remove the high frequency components, which may correspond to edges in the feature. In this work, an exhaustive analysis of the pooling options (normal pooling and Gaussian pooling (blur pool)) for classification, segmentation and autoencoders is presented. Two novel pooling approaches are presented such Laplacian-Gaussian Concatenation with Attention (LGCA) pooling and Wavelet based approximate-detailed coefficient concatenation with attention (WADCA) pooling. The details of each of the algorithms are explained below

### 2.1 Review of squeeze and excitation network

The fundamental operation in the proposed pooling layer is a channel attention network which combines the low frequency and high frequency feature maps. Let an image with resolution $(N_r \times N_c)$ be represented by $I \in \Re^{N_r \times N_c \times 3}$, where 3 represents the color channels. A neural network architecture (typically convolutional neural network (CNN) for images), takes the input image $I$ and extract features $F(I) \in \Re^{N_r \times N_c \times N_{ch}}$. These features $F(I)$ are passed through different layers bringing down the spatial resolution through pooling operations. As discussed, different types of pooling are developed to ensure the low and high frequency components are preserved for each of the feature layers. Hence both the frequency components are concatenated and passed through an attention layer in this work.

The squeeze and excitation (SE) network [18] have been employed for channel attention (as illustrated in Fig. 1) in the proposed approaches. The SE architecture has three parts, 1) *squeeze* operation which takes the features (input tensor) and aggregates the feature maps across their dimensions ($N_r \times N_c$) by producing an embedding, 2) *excitation* operation which takes the squeezed embedding as the input and estimates per-channel attention weights and 3) *scaling* where the actual weights are applied to the features to have attention weighted features. The SE architecture can be summarized as in (1).

$$SE\{F(X)\} = F_{attn} \times F(X) \tag{1}$$

where $F_{attn}$ are the attention weights of size $1 \times 1 \times N_{ch}$ and $X$ is the concatenated feature map ($X = I$ at the input stage). In the case of the proposed LGCA and WADCA approaches, the SE architecture ensures that channels of more importance are automatically attended to by the bigger network.

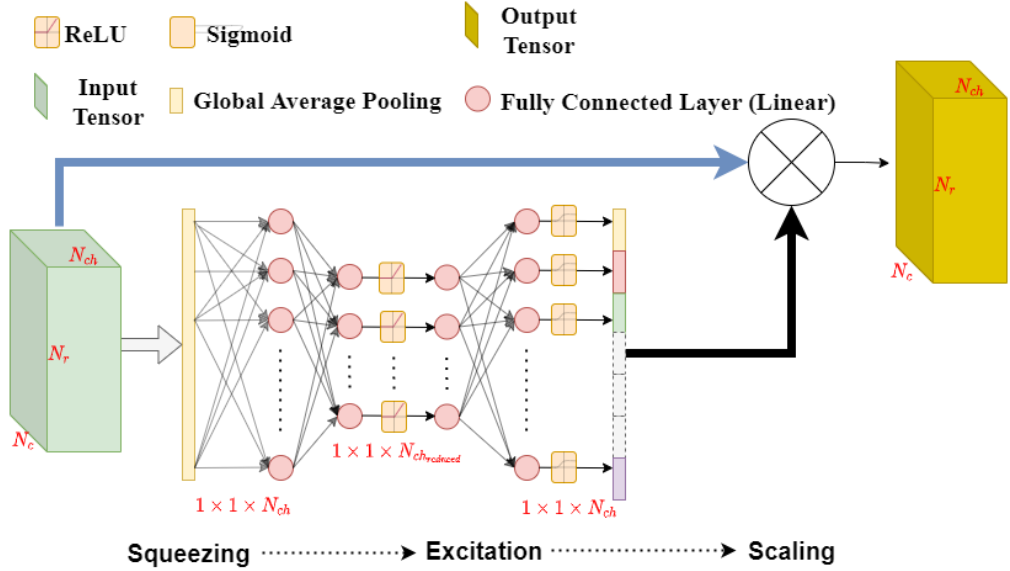

**Fig. 1 Channel attention implemented using squeeze and excitation Network as in [18]**

## 2.2 Laplacian-Gaussian Concatenation with Attention (LGCA) pooling

In the LGCA pooling approach, Gaussian and Laplacian filtering operations are performed on the input feature maps and are concatenated and passed through an attention network (shown in Fig. 1). Since the channel dimension is doubled during the concatenation operation, the output from the attention framework is passed through a convolution layer to reduce to the original dimension as the feature map. As discussed before, the intention of employing an attention network is to remove the redundancy among channels and to bring the focus of the overall network to the most relevant channels in the concatenated output while ensuring anti-aliasing of feature maps. The architecture for the LGCA approach is as shown in Fig. 2.

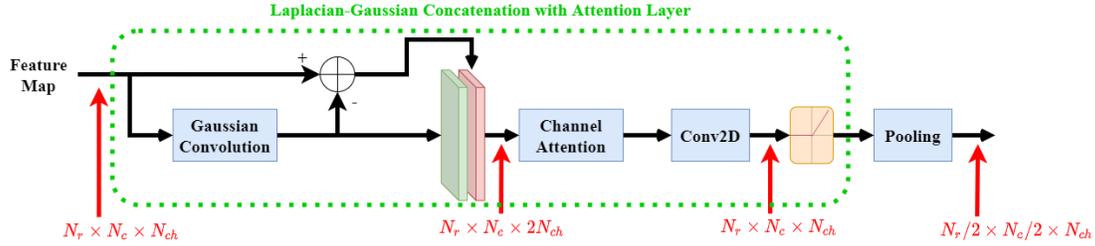

**Fig. 2 Proposed LGCA pooling framework**

The concatenation of the LGCA features is as in (2)

$$C\{F(X)\} = \{G(X), F(X) - G(X)\} \tag{2}$$

where $G(X)$ and $F(X) - G(X)$ are the Gaussian and Laplacian filtered components respectively. The Guassian Kernel values are adapted from [1].

## 2.3 Wavelet based approximate-detailed coefficient concatenation with attention (WADCA) pooling

Wavelets have been popular for model based denoising [19] and recently employed for down sampling in CNNS [2]. In the WADCA approach, 2D DWT based on Haar wavelet has been employed for the decomposition of the low and high frequency components. It has to be noted that, we have only employed one level of decomposition in the proposed approach. A detailed analysis of various wavelet bases and multiple levels of decomposition will be taken as a subsequent work. When compared to the LGCA approach, perfect quadrature mirror frequency splitting will be achieved and could always ensure the anti-aliasing pooling. The architecture for the WADCA approach is as shown in Fig. 3. The concatenation of the WADCA features is as in (3)

$$C\{F(X)\} = \{IDWT(\{zeros, W_a(X)\}), IDWT(\{zeros, W_d(X)\})\} \quad (3)$$

where $W_a(X)$ and $W_d(X)$ are the approximate and detailed wavelet coefficients respectively. It has to be noted that, $W_d(X)$ contains the horizontal, vertical and diagonal details as channels. In order to transform back to the spatial domain from the wavelet domain, 2D-IDWT is employed with concatenating the approximate and detailed coefficient with zeros. This is equivalent to approximate or detailed only reconstruction in the conventional case.

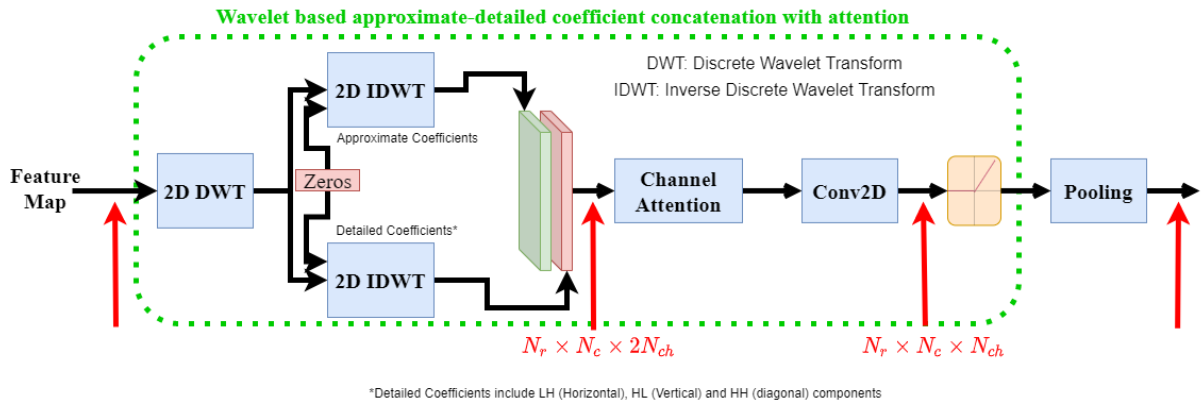

**Fig. 3 Proposed WADCA pooling framework**

    a. Proposed pooling approaches

In either of LGCA or WADCA, once the channel attention is done employing the SE architecture, the size of the channel will still be $N_r \times N_c \times 2N_{ch}$ due to the concatenation of low and high frequency components. However, to enable the possibility of including the proposed pooling layer in any established neural network architecture, it is desired to bring the dimensions back to the original feature dimensions of $N_r \times N_c \times N_{ch}$. To do this, we have proposed a convolutional layer followed by ReLU so that the aggregation of the channel dimension is decided by the application (e.g., classification or segmentation). This is followed by the desired pooling operation (e.g., average pooling, max pooling, strided convolution etc.).

## 3. Experiments

The proposed pooling architectures has been employed for various ablation studies by employing the same in various established architectures for the tasks of classification (e.g., ResNet50 [27], DenseNet121 [20] and MobileNetv2 [21]) and segmentation (e.g., SegNet [22] and U-Net [23]). The approach has also been included as part of the standard convolutional autoencoder (CAE) and observed the significance. The details of various ablation studies and their results with a brief explanation of the datasets are presented in the subsequent sections.

### 3.1 Datasets

Since the objective has been to prove the efficacy of the proposed pooling, the below datasets have been employed.

### 3.2 Cats and Dogs Dataset

This refers to a dataset presented in [24], which was used to develop the CAPTCHA that asks users to identify cats out of a set of 12 photographs of both cats and dogs. The dataset has 25,004 labelled photos, of which 12,502 are of cats and dogs each. The train-test split is 75:25 i.e 18753 training images and 6251 test images. This dataset was used to train and evaluate DenseNet121, MobileNetv2, ResNet50 and the custom convolutional autoencoder.

### 3.3 CamVid Dataset

The Cambridge-driving Labeled Video Database (CamVid) [25, 26] consists of five video sequences, captured using a camera mounted on the dash of a car. The five video sequences had a cumulative 701 frames. The database also contains ground truth labels that associate each pixel with one of 32 semantic classes. The dataset is split up as follows: 367 training pairs, 101 validation pairs and 233 test pairs. This dataset was used to train and evaluate the SegNet and UNet models

**Table I Details of the network trained**

| Application | Backbone | Epochs | Batch Size | Learning Rate (Reduce on Plateau) | Augmentations | Optimizer | Loss |
|---|---|---|---|---|---|---|---|
| Classifier | ResNet50 | 100 | 16 | 0.001 | Resize (256x256), CenterCrop(224x224), Normalize | Adam | Cross Entropy Loss |
| | DenseNet121 | 100 | 16 | 0.001 | Resize (256x256), CenterCrop(224x224), Normalize | SGD (momentum = 0.9, weight decay =4e-05) | Cross Entropy Loss |
| | MobileNetv2 | 100 | 16 | 0.001 | Resize (256x256), CenterCrop(224x224), Normalize | SGD (momentum = 0.9, weight decay =4e-05) | Cross Entropy Loss |
| Autoencoder | Custom | 100 | 16 | 0.001 | Resize (256x256), CenterCrop(224x224), Normalize | Adam | MSE Loss |
| Segmentation | U-Net | 100 | 8 | 0.0001 | Resize (256x256), CenterCrop(224x224) | SGD (momentum=0.99) | Focal Loss (alpha=0.8, gamma=2) |
| | SegNet | 100 | 8 | 0.0001 | Resize(256x256), CenterCrop(224x224) | SGD (momentum=0.99) | Focal Loss (alpha=0.8, gamma=2) |

## 3.4 Implementation Details

The proposed approach was implemented in PyTorch with models trained on NVIDIA DGX A100 Graphical Processing Units as part of the Centre for Development of Advanced Computing (CDAC) PARAM Siddhi AI system. The hyperparameters associated with each of the ablation studies along with other important details of the training are listed in Table I. We have tried to reproduce the original specifications as reported in [1, 27, 20-23]. The standard augmentations (affine transformations) have been avoided to check for the inherent robustness of the trained network towards the translations and rotations along with noise robustness.

## 4 Results

The proposed pooling architectures have been employed instead of standard pooling in the typical applications of classification, segmentation and autoencoders. The results of the same are described in detail as follows.

### 4.1 Classifiers

In this section, the results of employing the proposed LGCA and WADCA based pooling in ResNet50 [27], DenseNet121 [20] and MobileNetv2 [21] are presented. For the performance comparison, the following measures are employed as in [1, 2].

#### 4.1.1 Training performance

The different classifier backbones with various pooling options have been trained for 100 epochs with a batch size of 16 and a learning rate of 0.001 (with the option of reducing by factor of 10 on plateau). More details on the parameters are available in Table I. Fig. 4 shows the performance of the approaches for the initial 15 epochs. This was considered to observe any faster convergence for the different algorithms. It is interesting to note from Fig. 4 that, except for ResNet50, the proposed LGCA and WADCA outperforms both normal and Gaussian pooling in terms of faster learning/convergence in higher accuracy. This is extremely important in the scenarios where the dataset is limited.

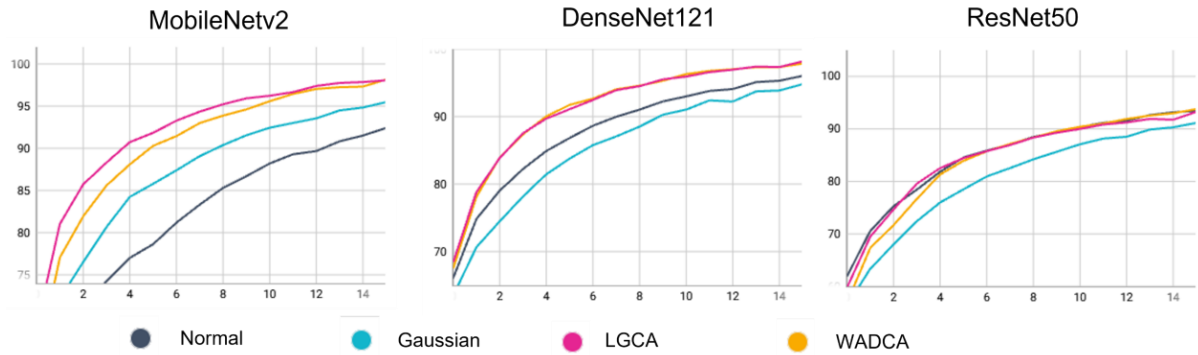

**Fig. 4 Performance comparison in terms of training accuracy convergence. The columns show results for MobileNetv2, DenseNet121, and ResNet50 respectively. Except for ResNet50, the proposed LGCA and WADCA outperforms both normal and Gaussian pooling in terms of faster learning/convergence in higher accuracy.**

### 4.1.2 Classification Accuracy with Rotation and Translation

One of the important aspects of the proposed pooling is the ability to preserve the high frequency components (edges). To show the effectiveness of the same, the accuracy on the test dataset of [24] for images which has been rotated up to 90 degrees and translated in random steps have been calculated. The results are averaged and reported in Table II. It is very clear from the results that the LGCA and the WADCA pooling approaches are certainly superior to the conventional pooling even in the normal augmentation of images, whereas for translation and rotation of images, the results are even more impressive with reduced mean and standard deviation.

**Table II Comparison of classification accuracy with rotation and translation of images**

| Classifier Backbone | Pooling Type | Accuracy (Normal) (Mean ± SD) % | Accuracy (Rotation) (Mean ± SD) % | Accuracy (Translation) (Mean ± SD) % |
|---|---|---|---|---|
| MobileNetv2 | Normal | 89.55 ± 8.11 | 77.38 ± 10.10 | 89.27 ± 8.14 |
| | Gaussian [1] | 92.38 ± 6.80 | 80.53 ± 9.47 | 90.94 ± 7.41 |
| | Proposed LGCA | 93.80 ± 6.48 | 82.94 ± 9.24 | 92.58 ± 6.54 |
| | Proposed WADCA | **94.29 ± 6.01** | **83.86 ± 9.08** | **93.81 ± 5.77** |
| DenseNet121 | Normal | 92.06 ± 6.75 | 90.36 ± 7.77 | 91.48 ± 6.97 |
| | Gaussian [1] | 91.58 ± 6.69 | 89.15 ± 8.04 | 90.39 ± 7.16 |
| | Proposed LGCA | **94.50 ± 5.83** | **93.57 ± 5.72** | **93.64 ± 6.46** |
| | Proposed WADCA | 94.37 ± 6.10 | 92.63 ± 7.12 | 92.84 ± 6.51 |
| ResNet50 | Normal | 92. 40 ± 6.69 | 82.32± 9.62 | 91.61 ± 6.98 |
| | Gaussian [1] | 92.73 ± 6.69 | 81.90 ± 10.36 | 92.01 ± 6.64 |
| | Proposed LGCA | **92.88 ± 6.41** | **82.53 ± 9.02** | **92.73 ± 6.96** |
| | Proposed WADCA | 92.09 ± 6.54 | 82.42 ± 9.37 | 91.62 ± 6.98 |

**4.1.3 Classification Consistency**

In [1], the performance has been evaluated using a measure known as classification consistency. This is basically done by rotation and translation of images by varying amounts and averaging the accuracy of predictions. In this work, the rotation is varied between 0 and 15 degrees in steps of 1 degree and translated between positions (0,0) and (12,12) of the top left corner. The results are averaged and reported in Table III. The superiority nature of the proposed LGCA and WADCA approaches are evident from Table III.

**Table III Comparison of classification consistency with rotation and translation of images**

| Classifier Backbone | Pooling Type | Accuracy (Normal) (Mean ± SD) % | Accuracy (Rotation) (Mean ± SD) % |
|---|---|---|---|
| MobileNetv2 | Normal | 82.90± 9.48 | 83.47 ± 9.79 |
| | Gaussian [1] | 86.29 ± 8.93 | 88.24 ± 8.12 |
| | Proposed LGCA | 88.09 ± 7.85 | 89.10 ± 7.55 |
| | Proposed WADCA | **90.35 ± 7.72** | **91.77 ± 7.25** |
| DenseNet121 | Normal | 87.23 ± 8.49 | 89.15 ± 7.73 |
| | Gaussian [1] | 86.01 ± 10.05 | 87.29 ± 9.25 |
| | Proposed LGCA | **91.35 ± 6.95** | **91.48 ± 7.05** |
| | Proposed WADCA | 90.27 ± 7.33 | 91.16 ± 7.50 |
| ResNet50 | Normal | **88.35 ± 8.50** | 87.92 ± 8.36 |
| | Gaussian [1] | 87.61 ± 8.27 | 88.94 ± 7.88 |
| | Proposed LGCA | 88.24 ± 8.43 | **89.83 ± 7.25** |
| | Proposed WADCA | 88.22± 8.14 | 89.01 ± 7.82 |

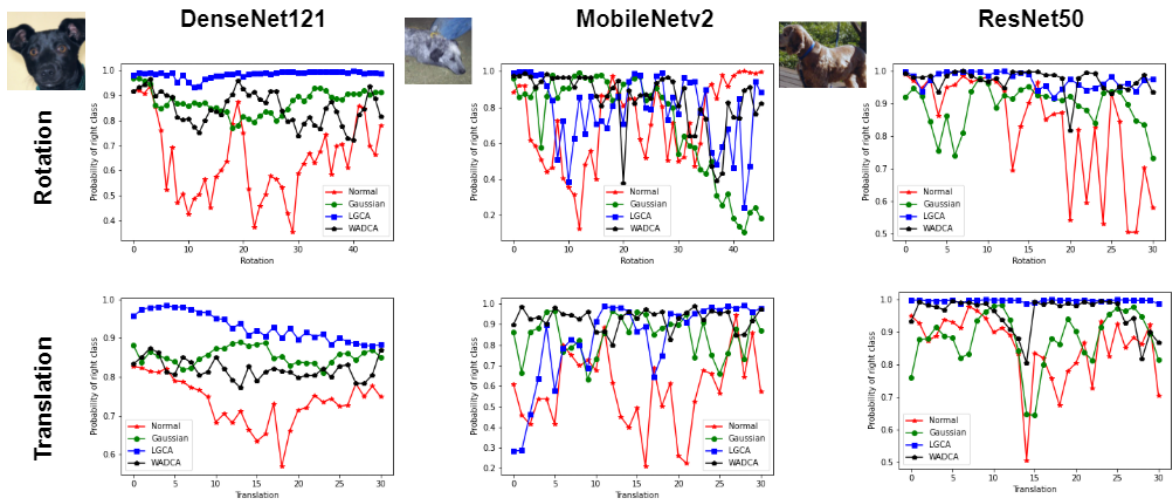

**Fig. 5 Performance comparison in terms of classification stability, where the probability of correct prediction is compared against controlled translations and rotations of a given image. First row shows results from the Rotation and Second row results from the Translation studies. The columns are for DenseNet121, MobileNetv2 and ResNet50 respectively. The results are mostly consistent for various images with some exceptions.**

### 4.1.4 Classification Stability

In this section, the classification stability [1], where the variation of probability of correct prediction is considered under systematic translation and rotation of images is considered. The results for the same are presented in Fig. 5. The columns correspond to a particular classifier backbone (DenseNet121, MobileNetv2 and ResNet50 respectively), whereas the rows correspond to rotations (first row) and translations (second row) of the images. It is very evident that the normal pooling has performance degradation (as reported in [1] also for translation cases) in correctly predicting the classes while translating and rotating the images. The LGCA approach appears to be more stable for DenseNet121 and ResNet50, whereas WADCA is more stable for MobileNetv2. The Gaussian approach in [1] (blurpool) is also doing a decent job and is more agnostic to rotations/translations when compared to normal pooling. Considering results presented in Table II and III and Fig. 5, it can be concluded that for DenseNet121 and ResNet50,

LGCA is doing better and for light weight networks like MobileNet, WADCA appears to perform better.

### 4.1.5 Noise robustness

In [1, 2, 28, 29], robustness to noise has been employed as a measure of the performance robustness. The idea is to corrupt the images with Gaussian noise and the same study has been redone in this section. The results have been analyzed using Grad-CAM [30] before and after the application of noise and are presented in Fig. 6. In all the cases, a mean zero additive Gaussian noise with standard deviation of 2 is applied. The important observations from Fig. 6 are that the normal pooling fails with noise, whereas the proposed LGCA and WADCA approaches are more robust to noisy observations. Also, it can be noticed that among the three classifiers, the ResNet50 is most robust to noise as evident from the results.

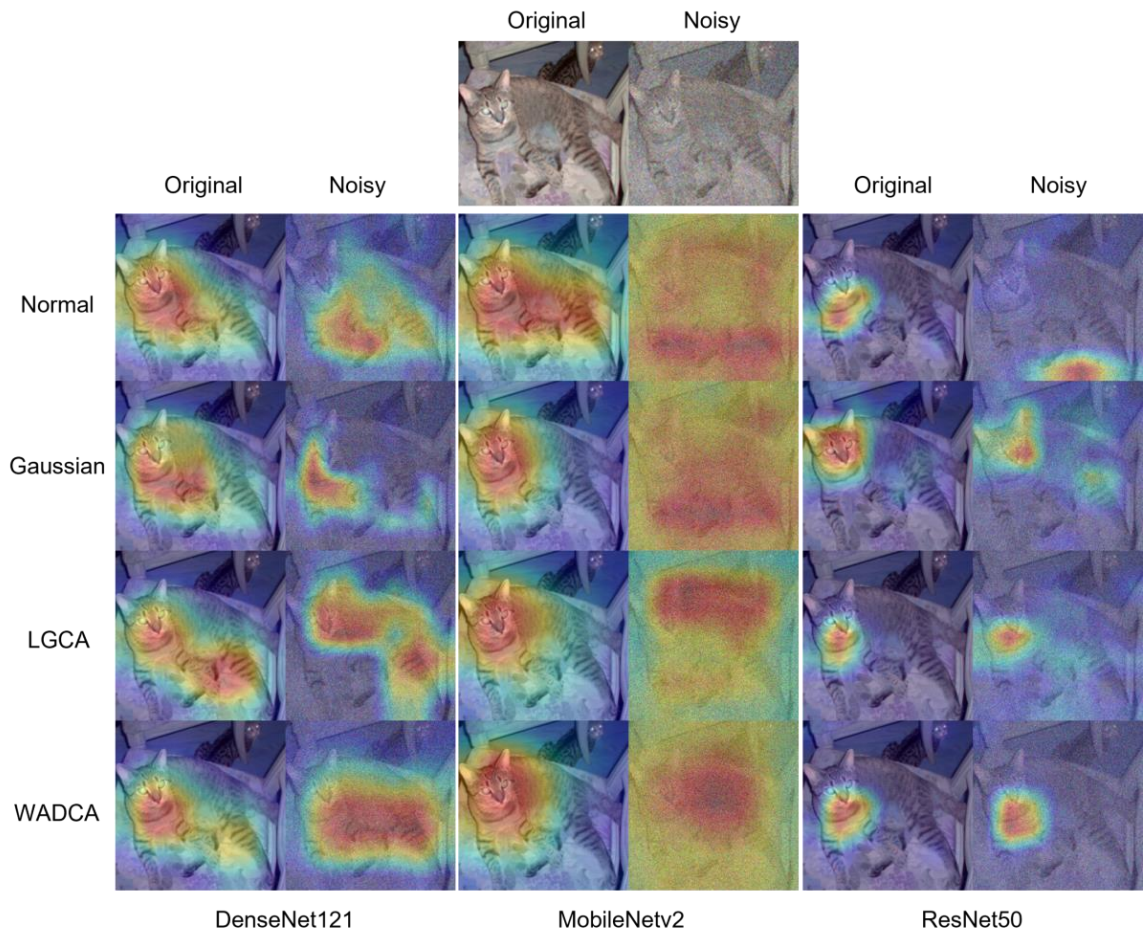

**Fig. 6 Performance comparison in terms of noise test for Normal (first row), Blur pooling (second row), proposed LGCA (third row) and proposed WADCA (fourth row) pooling schemes for DenseNet121 (columns 1 and 2), MobileNetv2 (columns 3 and 4) and ResNet50 (columns 5 and 6).**

### 4.1.6 t-distributed stochastic neighbor embedding (t-SNE) comparison

The t-SNE approach [31] has been employed to aid in dimensionality reduction for convenient and better visualization of features in the machine learning world. To compare the separability of the feature space, we have also employed the t-SNE for normal, rotation and translation cases and the results are presented in Fig. 7. It is very evident that the feature space is well separated in all the pooling architectures. However, when compared to normal pooling, the blur pooling and the proposed LGCA and WADCA are performing much better. In the translation and rotation cases, the proposed approaches seem to have better performance when compared to normal and blur pooling approaches. Another interesting observation is regarding highly non-linear separation for ResNet 50 features, and this may be due to possible over-fitting of the network given the binary classification problem.

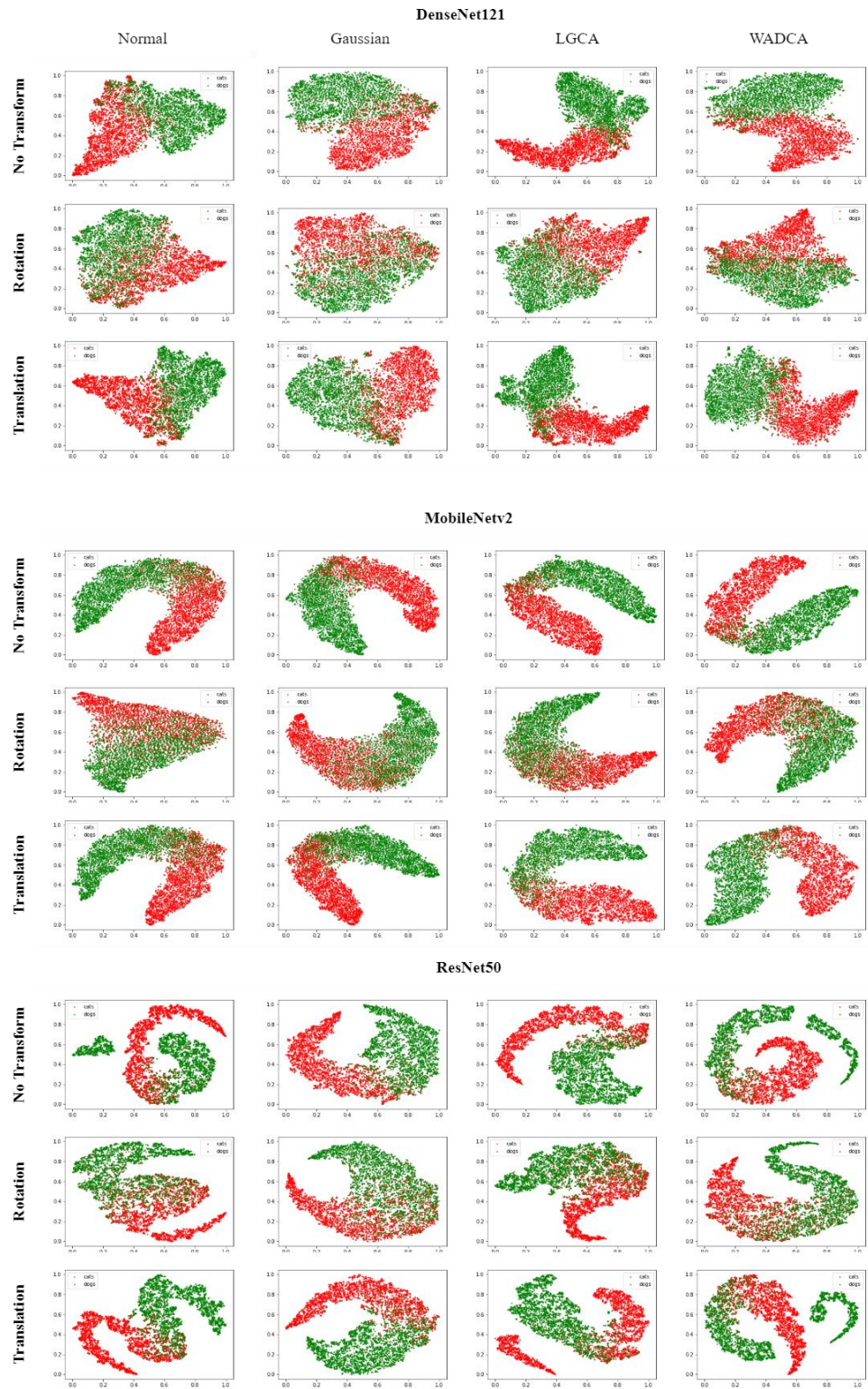

**Fig. 7 Performance comparison in terms of t-SNE for Normal (first column), Blur pooling (second column), proposed LGCA (third column) and proposed WADCA (fourth column) pooling schemes.**

**Table IV Comparison of IoU score with rotation and translation of images for segmentation architectures**

| Segmentation Architecture | Pooling Type | IoU Score (Normal) (Mean) | IoU Score (Rotation) (Mean ± SD) % | IoU Score (Translation) (Mean ± SD) % |
|---|---|---|---|---|
| U-Net | Normal | **0.81** | **0.237 ± 0.181** | **0.472 ± 0.138** |
| | Gaussian [1] | 0.792 | 0.238 ± 0.192 | 0.464 ± 0.161 |
| | Proposed LGCA | 0.753 | 0.227 ± 0.178 | 0.468 ± 0.145 |
| | Proposed WADCA | 0.803 | 0.187 ± 0.181 | 0.465 ± 0.141 |
| SegNet | Normal | 0.603 | 0.224 ± 0.187 | 0.449 ± 0.130 |
| | Gaussian [1] | 0.632 | 0.243 ± 0.182 | 0.477 ± 0.130 |
| | Proposed LGCA | 0.603 | 0.227 ± 0.173 | 0.462 ± 0.118 |
| | Proposed WADCA | **0.634** | **0.248 ± 0.184** | **0.488 ± 0.138** |

**4.2 Segmentation**

In this section, the results of employing the proposed LGCA and WADCA based pooling in segmentation networks such as U-Net [32] and SegNet [33] are presented. The intersection over union (IoU) scores for normal images, images rotated up to 90 degrees and translated in random steps are presented in Table IV. It is interesting to note that, for U-Net architecture, the normal pooling gives the best result which could be attributed to the effect of Residual connections. On the other hand, for a more regular encoder-decoder architecture like SegNet, the WADCA approach is having the best IoU score. The segmentation results for U-Net and SegNet architectures for various pooling options are shown in Figures 8 and 9. It is evident from the results that U-Net segmentation is better compared to SegNet as well as pooling is not as effective as in classifiers. However, if observed carefully, the enhancements in the sharpness of segmentation regions near edges (boundaries) for LGCA/WADCA pooling approaches are very evident from the results, which reiterates the importance of edge-preserved pooling in neural networks.

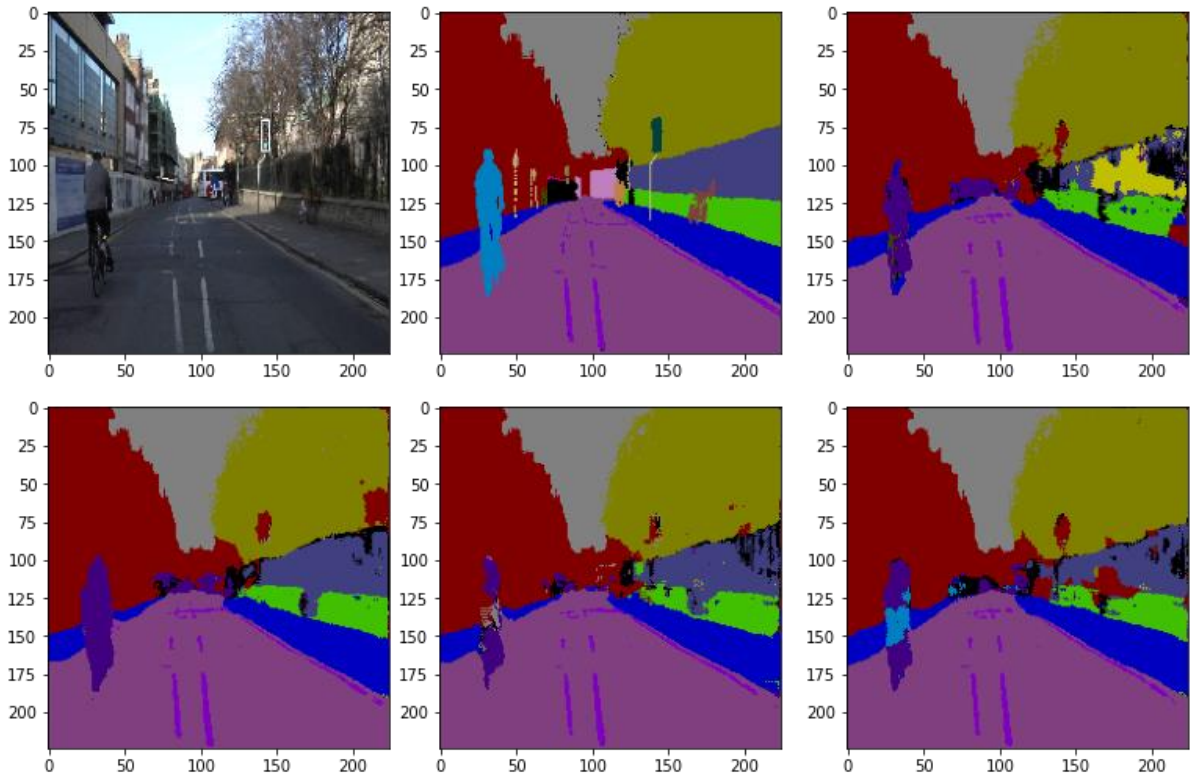

**Fig. 8 Comparison of U-Net results for various pooling schemes. Clockwise from top left : Original Image, Original Segmentation, Normal Pooling, Gaussian Pooling, LGCA pooling and WADCA pooling.**

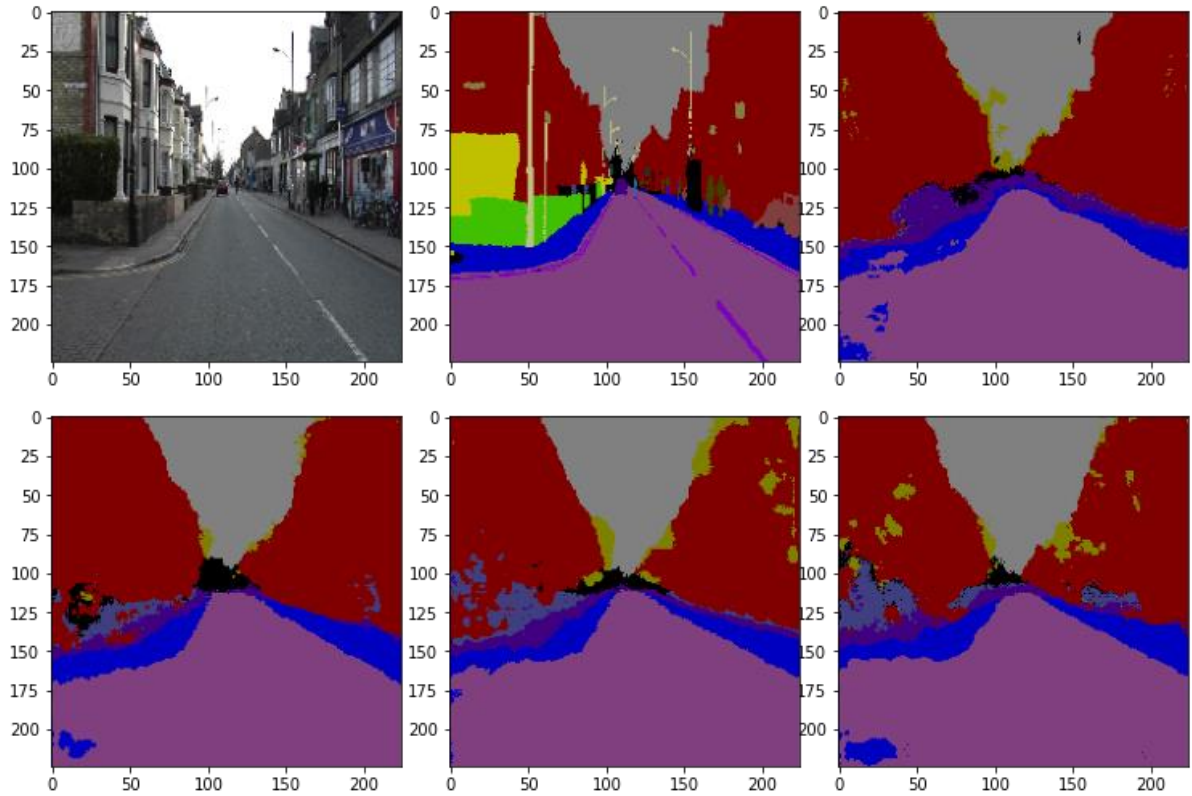

**Fig. 9 Comparison of SegNet results for various pooling schemes. Clockwise from top left : Original Image, Original Segmentation, Normal Pooling, Gaussian Pooling, LGCA pooling and WADCA pooling.**

### 4.3 Autoencoders

In this section, the results of employing the proposed LGCA and WADCA based pooling in a standard convolutional auto encoder setting are presented. The autoencoder architecture consists of four layers of encoder which reduces the channel dimension from 256x256 to 32x32 with channel dimensions as 48, 96, 192 and 32 respectively. The decoder as in any CAE is the transpose arrangement of the encoder. The MSE loss is multiplied by the batch size to avoid low error convergence issues. The MSE losses for normal images, images rotated up to 90 degrees and translated in random steps are presented in Table V.

**Table V Comparison of MSE with rotation and translation of images for CAE architecture**

| Pooling Type | MSE (Normal) (Mean) | MSE (Rotation) (Mean ± SD) % | MSE (Translation) (Mean ± SD) % |
|---|---|---|---|
| Normal | 10.77 | 16.158 ± 1.713 | 11.322 ± 1.631 |
| Gaussian [1] | 10.87 | 16.205 ± 1.722 | 11.365 ± 1.665 |
| Proposed LGCA | 10.76 | 16.161 ± 1.668 | 11.267 ± 1.561 |
| Proposed WADCA | 10.71 | 16.153 ± 1.631 | 11.292 ± 1.742 |

The reconstructed images for a sample case are also shown in Fig. 10. The results in Fig. 10 clearly show the significance of the edge preserving pooling. In the enlarged circles as shown in the second row in Fig. 10, the reconstruction particularly that of LGCA looks very close to the original image whereas for normal and Gaussian pooling there is a lot of smoothing as expected.

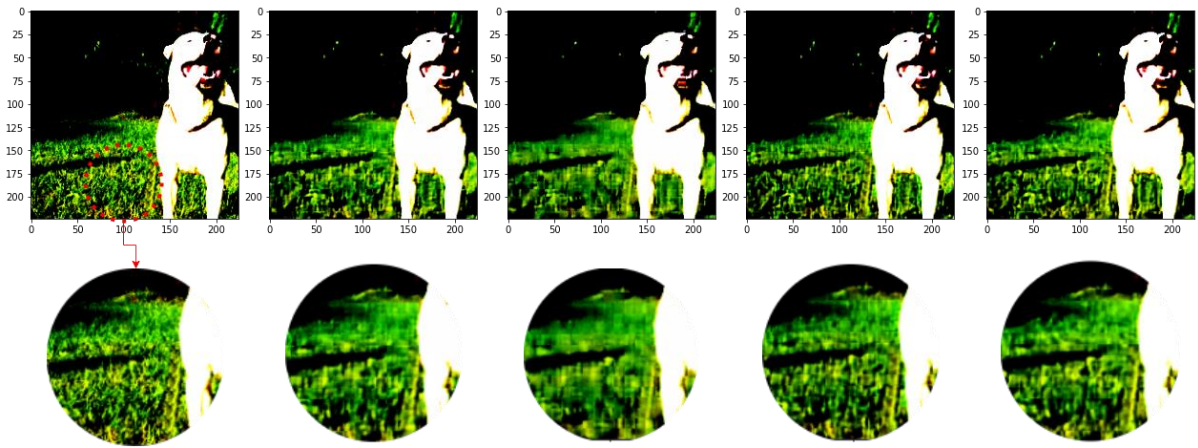

**Fig. 10 Comparison of CAE results for various pooling schemes. Left to right: Original Image, Reconstruction results using Normal Pooling, Gaussian Pooling, LGCA Pooling and WADCA Pooling. The red dotted circle is enlarged in the second row to show the significance of edge preserving pooling.**

## 5 Conclusions

The significance of pooling for shift invariance in convolutional neural networks has been attempted before. However, the fundamental building block of the anti-aliased CNN has been the application of Gaussian smoothing before the pooling operation, which will distort the edges in the feature maps. In this work, an exhaustive analysis of the edge preserving pooling options for classification, segmentation and convolutional autoencoders are presented. Two novel pooling approaches are presented such as Laplacian-Gaussian Concatenation with Attention (LGCA) pooling and Wavelet based approximate-detailed coefficient concatenation with attention (WADCA) pooling. The results suggest that the proposed pooling approaches outperform the conventional pooling as well as blur pooling for classification and convolutional autoencoders. In terms of average binary classification accuracy (cats vs dogs), the proposed LGCA and WADCA approaches outperforms the normal pooling and blur pooling. It is also observed from results that edge preserving pooling doesn't have any significance in segmentation tasks possibly due to high resolution to low resolution translation, whereas for convolutional auto encoders high resolution reconstruction has been observed for the LGCA pooling.

**Acknowledgements**

This work was supported by the Science and Engineering Research Board, Department of Science and Technology, Government of India, Grant No. CVD/2020/000221. We thank NVIDIA and C-DAC management for the successful completion of the project. The help provided by Mr.

Jigar Halani, Dr. Manish Modani, Mr. Megh Makwana and Mr. Prakash Tubakad (all from NVIDIA) for GPU access are thankfully acknowledged. The model training was carried out using the National PARAM Supercomputing Facility of C-DAC (PARAM SIDDHI AI system).

**Code Availability**

The codes for the proposed LGCA and WADCA pooling approaches are available for easy reference and usage at https://github.com/TheDarKnight13/Edge-Preserved-Universal-Pooling.